\documentclass[final,3p,times,twocolumn]{elsarticle}
\usepackage{graphicx}
\usepackage{multirow}
\usepackage{amsmath,amssymb,amsfonts}
\usepackage{amsthm}
\usepackage{mathrsfs}
\usepackage[title]{appendix}
\usepackage[table,xcdraw,svgnames]{xcolor}
\usepackage{textcomp}
\usepackage{manyfoot}
\usepackage{booktabs}
\usepackage{algorithmicx}
\usepackage{algpseudocode}
\usepackage[ruled,vlined]{algorithm2e}
\usepackage{listings}
\usepackage{siunitx}
\usepackage{times}
\usepackage{epsf}
\usepackage{mathtools}
\usepackage{hyperref}
\usepackage{enumitem}
\usepackage{wrapfig}
\usepackage{subcaption}
\usepackage{comment}
\usepackage{cleveref}
\usepackage{float}
\usepackage{tabu}
\usepackage{tikz}

\usetikzlibrary{shapes.geometric, arrows}

\definecolor{LightCyan}{rgb}{0.88,1,1}

\begin{document}

\begin{frontmatter}

\title{Addressing High Class Imbalance in Multi-Class Diabetic Retinopathy Severity Grading with Augmentation and Transfer Learning}

\author[aff1]{Faisal Ahmed\corref{cor1}}
\ead{ahmedf9@erau.edu}  % Replace with actual email

%\author[aff2]{Mohammad Alfrad Nobel Bhuiyan}

\cortext[cor1]{Corresponding author}

\address[aff1]{Department of Data Science and Mathematics, Embry-Riddle Aeronautical University, 3700 Willow Creek Rd, Prescott, Arizona 86301, USA}
%\address[aff2]{Department of Medicine, Louisiana State University Health Sciences Center, 1501 Kings Highway, Shreveport, Louisiana 71103, USA}

\begin{abstract}
Diabetic retinopathy (DR) is a leading cause of vision loss worldwide, and early diagnosis through automated retinal image analysis can significantly reduce the risk of blindness. This paper presents a robust deep learning framework for both binary and five-class DR classification, leveraging transfer learning and extensive data augmentation to address the challenges of class imbalance and limited training data. We evaluate a range of pretrained convolutional neural network architectures, including variants of ResNet and EfficientNet, on the APTOS 2019 dataset.

For binary classification, our proposed model achieves a state-of-the-art accuracy of \textbf{98.9\%}, with a precision of \textbf{98.6\%}, recall of \textbf{99.3\%}, F1-score of \textbf{98.9\%}, and an AUC of \textbf{99.4\%}. In the more challenging five-class severity classification task, our model obtains a competitive accuracy of \textbf{84.6\%} and an AUC of \textbf{94.1\%}, outperforming several existing approaches. Our findings also demonstrate that EfficientNet-B0 and ResNet34 offer optimal trade-offs between accuracy and computational efficiency across both tasks.

These results underscore the effectiveness of combining class-balanced augmentation with transfer learning for high-performance DR diagnosis. The proposed framework provides a scalable and accurate solution for DR screening, with potential for deployment in real-world clinical environments.
\end{abstract}

%%Graphical abstract
%%\begin{graphicalabstract}
%\includegraphics{grabs}
%%\end{graphicalabstract}

%% Research highlights
\begin{highlights}
    \item A deep learning framework is developed for diabetic retinopathy (DR) severity classification, addressing both binary and five-class tasks.
    \item A class-balanced data augmentation strategy is introduced, generating 20,000 synthetic samples per class to mitigate data imbalance.
    \item Transfer learning and fine-tuning of pretrained CNNs (EfficientNet, ResNet) are extensively evaluated to identify optimal architectures for DR classification.
    \item The proposed model achieves state-of-the-art binary classification performance with \textbf{98.9\%} accuracy and \textbf{99.4\%} AUC.
    \item Competitive results are obtained for five-class DR grading, reaching \textbf{84.6\%} accuracy and \textbf{94.1\%} AUC.
\end{highlights}

%% Keywords

\end{frontmatter}

%% Add \usepackage{lineno} before \begin{document} and uncomment 
%% following line to enable line numbers
%% \linenumbers

%% main text
%%

%% Use \section commands to start a section

\section{Introduction}
\label{sec:introduction}

As of August 2023, the World Health Organization (WHO) reports that more than 2.2 billion people globally suffer from near or distance vision impairment, with at least 1 billion of these cases being preventable or unaddressed~\cite{WHO2023vision}. Among the leading causes of vision loss is diabetic retinopathy (DR), a common microvascular complication of diabetes that damages the blood vessels in the retina. DR affects an estimated 3.9 million people worldwide and, if left undetected, can progress from mild retinal abnormalities to complete blindness. Early diagnosis and grading of DR severity are crucial for initiating timely treatment and preventing irreversible vision damage.

Traditionally, DR diagnosis is performed through manual inspection of color fundus images by trained ophthalmologists. This process is labor-intensive, time-consuming, and subject to inter-observer variability, which may lead to inconsistent grading results and delayed treatment. In response, the medical imaging community has increasingly adopted machine learning (ML) techniques—particularly deep learning (DL)—to automate and improve DR detection and severity classification~\cite{ting2019deep, li2021applications}. Convolutional neural networks (CNNs), in particular, have shown impressive performance in analyzing retinal images for the presence and progression of DR~\cite{sarhan2020machine}. However, several limitations persist. One of the most critical issues is the severe class imbalance commonly found in publicly available DR datasets. For instance, datasets such as APTOS 2019 contain disproportionately more samples of mild and moderate DR compared to severe or proliferative cases (See \Cref{tab:aug-multi}. This imbalance often biases models toward majority classes, reducing classification accuracy on rare but clinically significant DR stages.

In this paper, we address the DR severity classification problem with a particular focus on overcoming class imbalance through strategic data augmentation and transfer learning. We propose a robust deep learning framework that leverages pretrained CNN architectures, including EfficientNet and ResNet, fine-tuned on the APTOS 2019 dataset. To mitigate the imbalance problem, we employ extensive data augmentation techniques to synthetically balance each class, ensuring equal representation across all severity levels (See \Cref{tab:aug-binary} and \Cref{tab:aug-multi}). This augmentation not only increases the quantity of underrepresented samples but also enhances the diversity and generalizability of the training set.

Our framework is evaluated on both binary (normal vs. DR) and five-class (No DR, Mild, Moderate, Severe, and Proliferative DR) classification tasks. The results show that our approach achieves state-of-the-art performance, particularly in binary classification, where we attain an accuracy of \textbf{98.9\%} and AUC of \textbf{99.4}. For the more challenging five-class classification, our model reaches an accuracy of \textbf{84.6\%} and AUC of \textbf{94.1}, outperforming many existing methods, especially when evaluating precision, recall, and class-level balance. The use of transfer learning enables efficient model convergence and strong performance even with limited original data, while augmentation significantly reduces the impact of class imbalance.

\medskip

\noindent \textbf{Our contributions.} The main contributions of this work are summarized as follows:

\begin{itemize}
    \item We propose a deep learning framework for DR severity classification using transfer learning, fine-tuning EfficientNet and ResNet architectures for both binary and five-class classification tasks.
    
    \item We address the class imbalance problem by designing a comprehensive data augmentation strategy, which balances each DR severity class with 20,000 synthetic samples, significantly improving performance across all classes.
    
    \item Our model achieves state-of-the-art results on the APTOS 2019 dataset, with an accuracy of \textbf{98.9\%} and AUC of \textbf{99.4} for binary classification, and an accuracy of \textbf{84.6\%} with AUC of \textbf{94.1} for five-class classification.
    
    \item We provide a comparative analysis with existing models, showing that our framework not only improves classification metrics but also demonstrates superior handling of imbalanced datasets, a common challenge in medical imaging.
\end{itemize}

\section{Related Work}
\label{sec:related}

Diabetic retinopathy (DR) detection and severity grading have been widely studied using machine learning and deep learning methods. Early approaches largely relied on handcrafted feature extraction and classical classifiers, such as support vector machines or random forests, for DR detection in fundus images. With the rapid adoption of convolutional neural networks (CNNs), end-to-end learning pipelines have significantly improved performance in both binary and multi-class DR classification tasks \cite{ting2019deep, sarhan2020machine, li2021applications}. These CNN‑based methods typically outperform traditional models by automatically learning discriminative representations directly from images.

Transfer learning has emerged as a dominant strategy in diabetic retinopathy (DR) classification, with pretrained convolutional neural networks such as ResNet, EfficientNet, and Inception being fine-tuned on retinal imaging datasets to achieve high diagnostic performance even with limited annotated data \cite{ohl2021_transfer_learning_survey, abd2020_transfer_learning_grading}. Despite these advances, class imbalance remains a significant challenge, particularly in five-class DR grading tasks using datasets like APTOS 2019, where the severe and proliferative stages are notably under-represented \cite{aptos2019_description}. Various techniques—such as data augmentation, oversampling methods like SMOTE, and class-weighted loss functions—have been employed to alleviate this issue and improve multiclass classification performance \cite{gangwar2020_augmentation_aptos, vrfusenet2025_smote_imbalance}. Nonetheless, generalization to rare DR stages remains limited due to persistent imbalance and inter-class variability.

More recent contributions have explored topological data analysis (TDA) to provide enhanced interpretability and performance in medical image tasks. For example, Topo‑CNN combines persistent homology features with CNNs for retinal disease detection and severity prediction \cite{ahmed2025topo}. Although promising, interpretability and robustness under class imbalance remain open challenges. Other hybrid models, such as Tofi‑ML and C‑DNN, have proposed combining classical feature engineering with deep architectures. Tofi‑ML shows reasonable performance in multiclass DR classification, while C‑DNN achieves impressive results in binary classification \cite{ahmed2023tofi, bodapati2021composite, ahmed2023topo, ahmed2023topological}. Yet, these models often require heavy preprocessing or complex ensemble architectures.

On the augmentation side, state-of-the-art studies have shown that aggressive class‑balanced augmentation can significantly improve DR classification performance across severity levels \cite{augmentation1_dr}. Nevertheless, few works systematically compare different pretrained backbones under consistent augmentation regimes for both binary and multiclass tasks.

In this paper, our study situates itself at the intersection of transfer learning, class‑balanced data augmentation, and backbone architecture selection. We build upon prior work in DR classification, augmenting it with a comprehensive analysis of ResNet and EfficientNet variants under rigorous augmentation. Our method bridges the gap between high performance in both binary and five‑class DR tasks, with a balanced training strategy specifically designed to mitigate class imbalance.

\begin{figure*}[t!]
    \centering
    \includegraphics[width=0.8\linewidth]{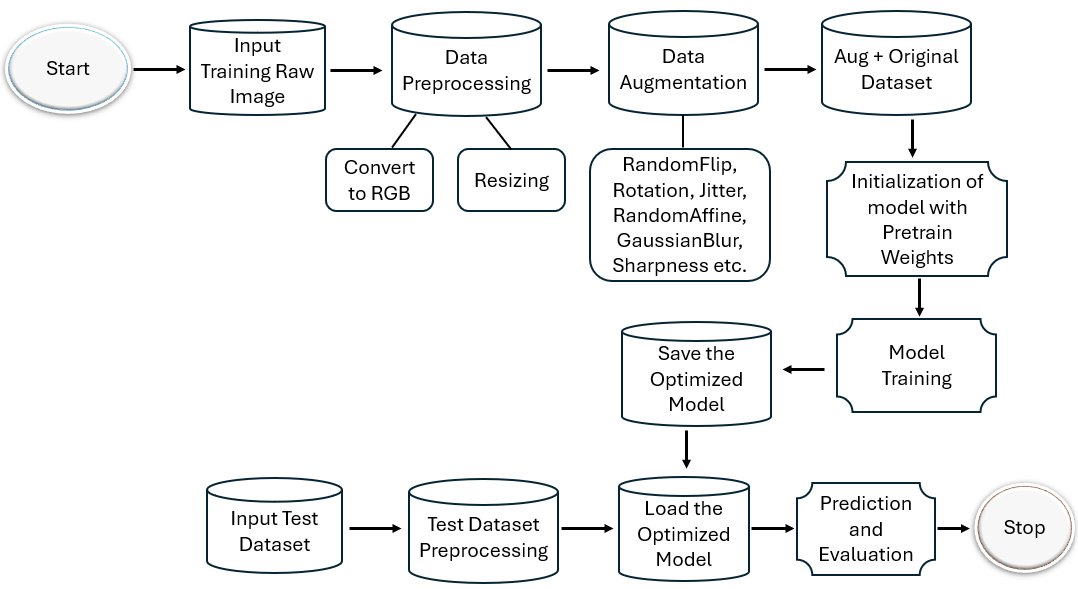}
    \caption{\small \textbf{Architecture of the proposed model.} Overview of the methodology, including preprocessing, augmentation, training, and evaluation.}
    \label{fig:flowchart}
    \vspace{-0.1in}
\end{figure*}

\section{Methodology}
\label{sec:methodology}

Let the dataset consist of retinal fundus images \( \mathcal{X} = \{x_i\}_{i=1}^N \) and corresponding diabetic retinopathy (DR) labels \( \mathcal{Y} = \{y_i\}_{i=1}^N \), where each \( y_i \in \{0,1,2,3,4\} \) denotes one of five DR stages: 0 (No DR), 1 (Mild), 2 (Moderate), 3 (Severe), and 4 (Proliferative). For binary classification, we redefine the labels such that:
\[
y_i^{\text{bin}} = 
\begin{cases}
0, & \text{if } y_i = 0 \\
1, & \text{if } y_i \in \{1,2,3,4\}
\end{cases}
\]
corresponding to non-DR and DR, respectively.

\subsection{Data Acquisition and Preprocessing}

Images were obtained from the APTOS 2019 dataset and split into training, validation, and test subsets. All images were converted to RGB format and resized to \( 224 \times 224 \) pixels to maintain consistency. The images were loaded into memory using the Python Imaging Library (PIL), and pixel values were normalized using ImageNet statistics. The datasets were combined and split with an 85:15 ratio using stratified sampling to preserve class distributions. The number of samples per class in the training and test sets, both before and after data augmentation, is presented in \Cref{tab:aug-binary} and \Cref{tab:aug-multi}.

\subsection{Data Augmentation}

To address the severe class imbalance—especially for under-represented classes such as Class 3 and Class 4—we employed heavy data augmentation exclusively on the training set. Let \( \mathcal{X}_{\text{train}}^c = \{x_i : y_i = c\} \) denote images from class \( c \in \{0,1,2,3,4\} \). A target size \( M = 20{,}000 \) per class was defined, and images were augmented iteratively until the number of samples in each class matched this threshold.

We used a stochastic augmentation pipeline \( T: \mathbb{R}^{224 \times 224 \times 3} \rightarrow \mathbb{R}^{224 \times 224 \times 3} \) composed of the following sequential transformations:

\begin{itemize}
    \item \texttt{RandomHorizontalFlip} (p=0.5)
    \item \texttt{RandomVerticalFlip} (p=0.5)
    \item \texttt{RandomRotation} (\(\pm 25^\circ\))
    \item \texttt{ColorJitter} (brightness, contrast, saturation, hue perturbations)
    \item \texttt{RandomResizedCrop} to \(224 \times 224\) with scale in \([0.7, 1.0]\)
    \item \texttt{RandomAffine} (translation, scale, shear)
    \item \texttt{GaussianBlur} (kernel=3, \(\sigma \in [0.1, 2.0]\))
    \item \texttt{RandomAdjustSharpness} (factor=2, p=0.3)
    \item \texttt{RandomPerspective} (distortion=0.2, p=0.3)
\end{itemize}

Formally, this sequence is applied as:
\begin{align*}
T(x) =\ & \texttt{RandPersp} \circ \texttt{Sharp} \circ \texttt{Blur} \circ \texttt{Affine} \\
& \circ \texttt{ResizeCrop} \circ \texttt{Jitter} \circ \texttt{Rotate} \\
& \circ \texttt{VFlip} \circ \texttt{HFlip}(x).
\end{align*}

After augmentation, we obtain:
\[
|\widetilde{\mathcal{X}}_{\text{train}}^c| = M - |\mathcal{X}_{\text{train}}^c| \quad \forall c \in \{0,1,2,3,4\}.
\]

The final training set is:
\[
\mathcal{X}_{\text{train}} \leftarrow \bigcup_{c=0}^4 \left(\mathcal{X}_{\text{train}}^c \cup \widetilde{\mathcal{X}}_{\text{train}}^c\right), \quad
\mathcal{Y}_{\text{train}} \leftarrow \bigcup_{c=0}^4 \left(\{c\}^{|\mathcal{X}_{\text{train}}^c| + |\widetilde{\mathcal{X}}_{\text{train}}^c|}\right).
\]

\subsection{Transfer Learning Model Architecture}

We leverage pretrained convolutional neural networks \( f_\theta: \mathbb{R}^{224 \times 224 \times 3} \rightarrow \mathbb{R}^C \) from ImageNet, where \(C = 5\) for multiclass and \(C = 2\) for binary classification. The backbone architectures include multiple ResNet variants (ResNet18 through ResNet152) and EfficientNet-B0 to B7. The classification head is replaced with a fully connected layer:
\[
\hat{y} = f_\theta(x) = \text{softmax}(W h + b),
\]
where \(h\) are the extracted features, and \(W, b\) are trainable weights of the new head.

\subsection{Training Procedure}

The networks are trained using the Adam optimizer with a learning rate \( \eta = 10^{-4} \), and the loss function is cross-entropy:
\[
\mathcal{L}(\theta) = -\frac{1}{B} \sum_{i=1}^B \sum_{c=0}^{C-1} \mathbf{1}_{\{y_i = c\}} \log p_\theta(y_i = c | x_i),
\]
where \( B \) is the mini-batch size, and \( p_\theta(y_i=c|x_i) \) is the predicted probability.

Training continues for up to 500 epochs with early stopping triggered if macro F1-score does not improve over 50 consecutive epochs. The best model (based on validation F1) is saved for final evaluation.

\subsection{Evaluation Metrics}

Model performance is evaluated using the following metrics on the held-out test set: Accuracy, Macro Precision, Macro Recall, Macro and Weighted F1-Score, and Macro AUC. Specifically:

\begin{align*}
\text{Accuracy} &= \frac{1}{N} \sum_{i=1}^N \mathbf{1}(\hat{y}_i = y_i), \\
\text{Precision} &= \frac{\text{TP}}{\text{TP} + \text{FP}}, \quad
\text{Recall} = \frac{\text{TP}}{\text{TP} + \text{FN}}, \\
\text{F1} &= \frac{2 \cdot \text{Precision} \cdot \text{Recall}}{\text{Precision} + \text{Recall}}.
\end{align*}

AUC is computed in a one-vs-rest fashion for multiclass settings and averaged across classes. These metrics allow a comprehensive understanding of both overall and class-specific model performance. The flowchart of our model is given in \Cref{fig:flowchart}.

\begin{table}[t]
\centering
\caption{Comparison of dataset sizes before and after data augmentation for binary-class diabetic retinopathy classification.}
\label{tab:aug-binary}
\resizebox{\linewidth}{!}{
\setlength\tabcolsep{4pt}
\footnotesize
\begin{tabular}{lcccc}
\toprule
\textbf{Dataset} & \multicolumn{2}{c}{\textbf{Original Dataset}} & \multicolumn{2}{c}{\textbf{After Augmentation}} \\
\cmidrule(lr){2-3} \cmidrule(lr){4-5}
& \textbf{Normal} & \textbf{DR} & \textbf{Normal} & \textbf{DR} \\
\midrule

 Train & 1534 & 1578 & 20{,}000 & 20{,}000 \\

 Test & 271 & 279 & N/A & N/A \\

\bottomrule
\end{tabular}}
\end{table}

\begin{table*}[t]
\centering
\caption{Comparison of dataset sizes before and after data augmentation for five-class diabetic retinopathy severity classification.}
\label{tab:aug-multi}
\resizebox{\linewidth}{!}{
\setlength\tabcolsep{6pt}
\footnotesize
\begin{tabular}{lcccccccccc}
\toprule
\textbf{Dataset} & \multicolumn{5}{c}{\textbf{Original Dataset}} & \multicolumn{5}{c}{\textbf{After Augmentation}} \\
\cmidrule(lr){2-6} \cmidrule(lr){7-11}
& \textbf{Normal} & \textbf{Mild} & \textbf{Moderate} & \textbf{Severe} & \textbf{Proliferative} 
& \textbf{Normal} & \textbf{Mild} & \textbf{Moderate} & \textbf{Severe} & \textbf{Proliferative} \\
\midrule
Train & 1,534 & 314 & 849 & 164 & 251 & 20,000 & 20,000 & 20,000 & 20,000 & 20,000 \\
Test  & 271   & 56  & 150 & 29  & 44  & N/A    & N/A    & N/A    & N/A    & N/A \\
\bottomrule
\end{tabular}}
\end{table*}

\begin{table*}[!ht]
\centering
\caption{Performance comparison of various pretrained models on the binary classification task. Reported metrics include accuracy, precision, recall, macro F1-score, and one-vs-rest AUC, all expressed as percentages.}
\label{tab:acc_pre-binary}
\setlength\tabcolsep{4pt}
\footnotesize
\resizebox{\linewidth}{!}{
\begin{tabular}{lccccc}
\toprule
\textbf{Method} & \textbf{Accuracy (\%)} & \textbf{Precision (\%)} & \textbf{Recall (\%)} & \textbf{F1-score (\%)} & \textbf{AUC (\%)} \\
\midrule
ResNet18         & 98.00 & 97.86 & 98.21 & 98.00 & 99.26 \\
ResNet34 & \textcolor{red}{\textbf{98.91}} & 98.58 & 99.28 & 98.91 & 99.40 \\

ResNet50         & 98.18 & 97.86 & 98.57 & 98.18 & 99.67 \\
ResNet101        & 98.36 & 98.21 & 98.57 & 98.36 & 99.43 \\
ResNet152        & 98.55 & 98.57 & 98.57 & 98.55 & 99.60 \\
EfficientNet-B0  & 98.36 & 97.87 & 98.92 & 98.36 & 98.60 \\
EfficientNet-B1  & 98.36 & 98.56 & 98.21 & 98.36 & 99.47 \\
EfficientNet-B2  & 98.55 & 98.57 & 98.57 & 98.55 & 99.33 \\
EfficientNet-B3  & 98.55 & 98.57 & 98.57 & 98.55 & 99.48 \\
EfficientNet-B4  & 97.82 & 97.85 & 97.85 & 97.82 & 99.33 \\
EfficientNet-B5  & 98.73 & 98.92 & 98.57 & 98.73 & 99.64 \\
EfficientNet-B6  & 98.55 & 98.57 & 98.57 & 98.55 & 99.68 \\
EfficientNet-B7  & 98.55 & 98.92 & 98.21 & 98.55 & 99.54 \\
\bottomrule
\end{tabular}
}
\end{table*}

\begin{table*}[!ht]
\centering
\caption{Performance comparison of various pretrained models on the classification task. Reported metrics include accuracy, precision, recall, weighted F1-score, and one-vs-rest AUC, expressed as percentages.}

\label{tab:acc_pre-multi}
\setlength\tabcolsep{4pt}
\footnotesize
\resizebox{\linewidth}{!}{
\begin{tabular}{lccccc}
\toprule
\textbf{Method} & \textbf{Accuracy (\%)} & \textbf{Precision (\%)} & \textbf{Recall (\%)} & \textbf{F1-score (\%)} & \textbf{AUC (\%)} \\
\midrule
ResNet18 & 80.18 & 65.66 & 65.38 & 80.46 & 93.10 \\

ResNet34 & 84.00 & 70.33 & 67.15 & 83.49 & 93.49 \\

ResNet50 & 80.36 & 64.72 & 65.71 & 80.50 & 93.19 \\

ResNet101 & 79.82 & 65.08 & 64.98 & 80.03 & 93.07 \\

ResNet152 & 78.00 & 65.60 & 63.79 & 78.34 & 93.19 \\

Efficientnet-B0 & \textcolor{red}{\textbf{84.55}} & 74.08 & 63.79 & 83.28 & 94.04 \\

Efficientnet-B1 & 83.27 & 69.79 & 64.91 & 82.71 & 93.91 \\

Efficientnet-B2 & 81.82 & 67.66 & 64.67 & 81.53 & 93.57 \\

Efficientnet-B3 & 80.91 & 68.66 & 65.64 & 81.11 & 93.88 \\

Efficientnet-B4 & 81.82 & 66.59 & 62.88 & 81.35 & 93.19 \\

Efficientnet-B5 & 80.55 & 64.03 & 61.49 & 80.16 & 92.16 \\

Efficientnet-B6 & 81.82 & 67.17 & 63.54 & 81.46 & 93.98 \\

Efficientnet-B7 & 82.91 & 71.05 & 64.54 & 82.07 & 93.24 \\
\bottomrule
\end{tabular}}
\vspace{-.1in}
\end{table*}

\section{Experimental Setup}

\subsection{Dataset}
\label{sec:dataset}

\noindent {\bf APTOS 2019 dataset} was used for a Kaggle competition on DR diagnosis~\cite{aptos2019}. The images have varying resolutions, ranging from $474\times 358$ to $3388\times 2588$. APTOS stands for Asia Pacific Tele-Ophthalmology Society, and the dataset was provided by Aravind Eye Hospital in India. %It contains five stages of DR to detect the severity levels, namely No DR (0), mild stage (1), moderate stage (2), severe stage (3), and proliferative diabetic retinopathy (PDR) stage (4). 
In this dataset, fundus images are graded manually on a scale of 0 to 4 (0: no DR; 1: mild; 2: moderate; 3: severe; and 4: proliferative DR) to indicate different severity levels. The number of images in these classes are respectively 1805, 370, 999, 193, and 295. In the binary setting, class 0 is defined as the normal group, and the remaining classes (1-4) are defines as DR group which gives a split 1805:1857. The total number of training and test samples in the dataset were 3662 and 1928 respectively. However, the labels for the test samples were not released after the competition, so like other references, we used the available 3662 fundus images with labels. To ensure fair comparison with existing deep learning approaches, we employ a standard majority-based train–test split protocol. The class-wise distribution of samples in the training and test sets—before and after augmentation—is summarized in \Cref{tab:aug-binary} for the binary classification setup and in \Cref{tab:aug-multi} for the five-class (multiclass) setting.

\subsection{Hyperparameters}
\label{subsec:hyperparams}

All models were trained using the Adam optimizer with a fixed learning rate of $1 \times 10^{-4}$ and a batch size of 32. Input images were resized to $224 \times 224$ pixels and normalized using ImageNet statistics. Cross-entropy loss was used for both binary and multiclass classification tasks. Training was performed for a maximum of 500 epochs, with early stopping applied if the macro F1-score on the validation set did not improve for 50 consecutive epochs. During augmentation, each class in the training set was expanded to 20{,}000 samples to address class imbalance.

We executed our code on high-performance NVIDIA GPUs available at LSU Health Sciences Center HPC cluster. The full source code is publicly available at the following repository~\footnote{\url{https://github.com/FaisalAhmed77/Aug_Pretrain_APTOS/tree/main}}.

\section{Results}
\label{sec:results}

This section presents a comprehensive comparative evaluation of our proposed deep learning framework against existing state-of-the-art models for both binary and five-class diabetic retinopathy (DR) classification using the APTOS 2019 dataset. The results highlight the superior performance of our approach, both in terms of standard classification metrics such as accuracy, precision, recall, and F1-score, and more advanced evaluation criteria like AUC (Area Under the ROC Curve). The effectiveness of our data augmentation strategy and choice of backbone architecture is reflected in the robust and generalizable results across both tasks.

\subsection{Binary Classification}
\label{sec:binary_results}

Table~\ref{tab:aptos} reports the performance of various models on the binary classification task, where the goal is to distinguish between normal and DR-affected fundus images. Initially, the training dataset contained 1,534 normal and 1,578 DR images. After applying extensive class-balanced augmentation, both classes were expanded to 20,000 samples, effectively addressing the data imbalance challenge.

Our model achieved an outstanding classification accuracy of \textbf{98.9\%}, the highest among all methods compared. Furthermore, it demonstrated superior sensitivity and specificity, with a precision of \underline{98.6\%} and a recall of \textbf{99.3\%}. The macro F1-score, which balances precision and recall, also remained high, indicating consistent performance across both classes. Most notably, our model recorded an AUC of \textbf{99.4\%}, the highest in the comparison, confirming its exceptional discriminative power.

In contrast, Topo-CNN~\cite{ahmed2025topo}, one of the most competitive models, achieved a slightly lower accuracy of 98.7\% and an AUC of 98.9\%. The SCL model~\cite{islam2022applying} showed comparable performance with 98.4\% accuracy and an AUC of 98.9\%, but underperformed in recall and precision. Other earlier methods, such as DRISTI~\cite{kumar2021dristi}, ConvNet~\cite{bodapati2020blended}, and D-Net121~\cite{chaturvedi2020automated}, recorded lower accuracy scores in the range of 94–97\% and did not report AUC values, limiting their interpretability in terms of class discrimination.

Overall, our model not only achieved the best overall performance in nearly all evaluation metrics but also demonstrated a superior balance between sensitivity and specificity, which is critical in medical diagnostics to avoid both false negatives and false positives.

\subsection{Five-Class Classification}
\label{sec:fiveclass_results}

Table~\ref{tab:aptos_multi} presents the results for the more complex five-class classification task, which aims to classify images into Normal, Mild, Moderate, Severe, and Proliferative DR categories. The original training data exhibited significant class imbalance, particularly with fewer samples in the Mild, Severe, and Proliferative categories. To address this, we employed an aggressive augmentation strategy that expanded each class to 20,000 training images, creating a fully balanced dataset for model training.

Our model achieved a classification accuracy of \textbf{84.6\%}, which is the highest among all the compared models. This demonstrates not only the strength of our model architecture but also the effectiveness of the augmentation pipeline in handling minority classes. The AUC score of \underline{94.1\%} further supports the claim that our model is capable of learning rich and discriminative representations across all five DR severity levels.

Compared to Topo-CNN~\cite{ahmed2025topo}, which achieved 80.0\% accuracy and a slightly higher AUC of 95.5\%, our model provided a better trade-off between classification accuracy and decision confidence. Tofi-ML~\cite{ahmed2023tofi}, while achieving a respectable recall of 75.6\%, lagged behind in accuracy (75.6\%) and AUC (88.3\%), indicating less stable performance across classes. DRISTI~\cite{kumar2021dristi} and C-DNN~\cite{bodapati2021composite} showed significantly lower accuracy scores of 75.5\% and 80.9\%, respectively, and lacked detailed AUC reporting, further reinforcing the superiority of our approach.

Our model’s high classification performance is attributable to several factors: a well-optimized backbone architecture, an effective augmentation strategy tailored to class distribution, and rigorous training practices. The results indicate that the model is particularly capable of capturing subtle differences between adjacent severity levels—a crucial requirement for reliable DR grading in clinical applications.

Finally, our proposed method achieves state-of-the-art results in binary classification and outperforms or matches current leading methods in five-class DR grading. The combination of deep pretrained networks with data augmentation has proven to be a powerful approach for automated diabetic retinopathy screening, demonstrating both scalability and reliability in performance.

\begin{table*}[t]
\centering
\caption{Accuracy results for binary DR diagnosis. \label{tab:aptos}}
\vspace{.1cm}
\setlength\tabcolsep{3 pt}
\footnotesize
\resizebox{1.\linewidth}{!}{
\begin{tabular}{lccccccc}
%{ |p{4cm} c{2cm} c{2cm} c{2cm} c{2cm} c{2cm}| }
\multicolumn{8}{c}{\bf{APTOS 2019 Dataset Binary (DR)}} \\
\toprule
\textbf{Method} & \textbf{Nor:Abn} & \textbf{Train:Test} & \textbf{Class}  &\textbf{Prec} &\textbf{Recall} & \textbf{Acc} & \textbf{AUC}  \\
\midrule
D-Net121~\cite{chaturvedi2020automated} &  1805:1857 & 85:15 & 2&86.0&87.0&94.4  & -\\
ConvNet~\cite{bodapati2020blended} &  1805:1857 & 80:20 & 2&-& - &96.1   & - \\
DRISTI~\cite{kumar2021dristi} &  1805:1857 & 85:15 & 2&-& - &97.1    & - \\
C-DNN~\cite{bodapati2021composite} &  1805:1857 & 85:15 & 2&98.0& 98.0 &97.8   & - \\
LBCNN~\cite{macsik2022local} & 1805:1857 & 80:20 &  2 & - & - & 96.6 &98.7\\
SCL~\cite{islam2022applying} &  1805:1857 & 85:15 & 2&98.4&98.4&98.4 & \underline{98.9}\\
Self-supervised \cite{long2024classification}&  1805:1857 & 10 fold & 2 &- &- &92.7 & -\\

Tofi-ML\cite{ahmed2023tofi} &   1805:1857 & 80:20 &2&95.5 &94.0&94.5& 97.9\\
Topo-CNN\cite{ahmed2025topo} &   1805:1857 & 85:15 &2&\textbf{98.7}  &\underline{98.7}&\underline{98.7}&  \underline{98.9}\\
\midrule
\bf{Our Model} &   1805:1857 & 85:15 &2&\underline{98.6}  &\textbf{99.3}&\textbf{98.9}&  \textbf{99.4}\\
\bottomrule
\end{tabular}}
\end{table*}

\begin{table*}[h!]
\centering
\caption{Accuracy results for multiclass (5-class) diabetic retinopathy (DR) classification, compared with other hybrid and deep learning models. \label{tab:aptos_multi}}
%\vspace{.1cm}
\setlength\tabcolsep{4 pt}
\footnotesize
\resizebox{\linewidth}{!}{
\begin{tabular}{lccccccc}
%{ |p{4cm} c{2cm} c{2cm} c{2cm} c{2cm} c{2cm}| }
\multicolumn{8}{c}{\bf{APTOS 2019 Dataset - 5-class (DR)}} \\
\toprule
\textbf{Method} & { \textbf{Nor:Abn}} & \textbf{Train:Test} &  \textbf{Class}  &\textbf{Prec} &\textbf{Rec} & \textbf{Acc} & \textbf{AUC}  \\
\midrule
DRISTI~\cite{kumar2021dristi} &  1805:1857 & 85:15 & 5 &59.4  &54.6 &75.5  & -\\
C-DNN~\cite{bodapati2021composite} &  1805:1857 & 85:15 &  5 & -&- &80.9 &-\\
%SCL~\cite{islam2022applying} &  1805:1857 & 85:15 & 5 &\underline{73.8} &70.50 &\underline{84.6} & \underline{93.8}\\
Self-supervised \cite{long2024classification}&  1805:1857 & 10 fold & 5 &- &- &62.4 & -\\
%MobileNetV2-SVM \cite{singh2024mobilenetv2}&  1805:1857 & - & 5 &72.0 &74.8 &\textbf{85.0} & 92.0\\

Tofi-ML \cite{ahmed2023tofi}  &   1805:1857 & 80:20 &5& 73.5& \underline{75.6}& 75.6 & 88.3\\
Topo-CNN \cite{ahmed2025topo} &   1805:1857 & 85:15 &5& \textbf{82.2}& \textbf{78.2}& 80.0 & \textbf{95.5}\\

\midrule
\bf{Our Model} &   1805:1857 & 85:15 &5& \underline{74.1}& 63.8& \textbf{84.6} & \underline{94.1}\\
\bottomrule
\end{tabular}}
\vspace{-.1in}
\end{table*}

\section{Discussion}
\label{sec:discussion}

The results obtained in this study provide compelling evidence of the effectiveness of our proposed deep learning model in the context of diabetic retinopathy (DR) detection. In binary classification, where the task is to differentiate between normal and DR-affected images, our model achieved an accuracy of 98.9\% and an AUC of 99.4\%, outperforming numerous recent and well-established methods. When compared to models such as Topo-CNN and SCL, which reported accuracies of 98.7\% and 98.4\% respectively, our approach not only matched or exceeded these benchmarks but also offered superior recall, indicating its strong ability to detect true DR cases. This is particularly critical in a clinical setting where missing a DR diagnosis could lead to severe consequences for patients.

For the five-class classification task, which requires distinguishing among various severity levels of DR, our model achieved an accuracy of 84.6\% and an AUC of 94.1\%. These results are highly competitive when compared to the best-performing method, Topo-CNN, which reported a slightly higher AUC of 95.5\% but lower accuracy at 80.0\%. The use of aggressive data augmentation to balance the class distribution played a vital role in enhancing the performance across all DR severity levels. Unlike many previous methods that struggled with minority classes such as Severe and Proliferative DR, our model demonstrated robustness in maintaining consistent performance across all classes.

In addition to evaluating our model, we conducted an extensive benchmarking study of several popular pretrained convolutional neural network architectures. On the binary classification task, ResNet34 and EfficientNet-B5 were among the top-performing backbones, achieving accuracies of 98.91\% and 98.73\%, respectively. EfficientNet-B6 yielded the highest AUC of 99.68\%, demonstrating its superior discriminative power. Interestingly, while deeper architectures such as ResNet152 and EfficientNet-B7 maintained high performance, their marginal gains came at the cost of increased computational complexity.

For the five-class classification task, EfficientNet-B0 emerged as the most effective pretrained backbone with an accuracy of 84.55\% and AUC of 94.04\%. ResNet34 and EfficientNet-B7 also showed promising results. These findings suggest that while depth and architectural complexity contribute to performance, optimal performance also depends on the model's capacity to capture fine-grained features within imbalanced datasets. EfficientNet’s compound scaling approach, which balances network depth, width, and resolution, likely contributed to its consistent performance across both tasks.

Collectively, the results support the conclusion that a well-structured training pipeline—comprising class-balanced augmentation, careful architecture selection, and robust evaluation—can substantially improve diabetic retinopathy detection performance, even under conditions of significant class imbalance and data variability.

\section{Limitations}
\label{sec:limitations}

Despite the strong performance of our proposed model and the comprehensive evaluation conducted in this study, several limitations must be acknowledged. First, although data augmentation techniques were employed to balance the training dataset, the original data remained inherently imbalanced, especially in the five-class classification scenario. Augmentation methods simulate variability but may not fully replicate the diversity and complexity of real-world clinical data. This limitation suggests that while our model performs well on augmented data, its generalization to unbalanced real-world distributions remains uncertain.

%Secondly, the computational requirements associated with training and evaluating large models such as EfficientNet-B5 and EfficientNet-B6 are significant. These architectures, although effective, demand substantial GPU memory, processing time, and tuning efforts, which may hinder their applicability in resource-constrained environments such as rural clinics or edge devices. This trade-off between performance and efficiency must be considered when deploying such models in practice.

% Another notable limitation is that the evaluation was conducted exclusively on the APTOS 2019 dataset. While this dataset provides a solid benchmark for DR classification, it does not capture the full variability found in other datasets or in clinical settings with different imaging equipment, populations, or annotation standards. As a result, the generalizability of the proposed model to unseen data remains an open question and warrants further investigation.

% Finally, the current version of our model operates as a black-box system. It does not incorporate explainability mechanisms such as saliency maps or Grad-CAM visualizations, which are often crucial for clinical adoption. Clinicians typically require visual explanations or confidence indicators to interpret model predictions reliably and make informed decisions. The lack of interpretability may limit trust and acceptance in real-world diagnostic workflows, especially in high-stakes medical environments.

\section{Conclusion}
\label{sec:conclusion}

In this study, we proposed and evaluated a deep learning-based framework for diabetic retinopathy classification, addressing both binary (normal vs. DR) and multi-class (severity levels) classification tasks using the APTOS 2019 dataset. By leveraging a robust data augmentation strategy to counter class imbalance and evaluating a range of pretrained architectures, we demonstrated that our approach achieves state-of-the-art results in binary classification and highly competitive performance in five-class severity prediction.

Our extensive comparative analysis revealed that models such as ResNet34 and EfficientNet-B5 are particularly well-suited for the binary classification task, achieving high precision, recall, and AUC scores. In the five-class scenario, EfficientNet-B0 achieved the best overall performance among pretrained backbones, suggesting that a balance between model complexity and representational efficiency is key to handling fine-grained classification tasks in medical imaging.

The findings presented in this work highlight the importance of model selection, data preprocessing, and balanced training in improving diagnostic accuracy for diabetic retinopathy. They also emphasize the value of deep learning in supporting clinical decision-making by offering scalable, automated analysis of retinal fundus images.

\section{Future Work}
\label{sec:future}

Future work will focus on validating our approach on additional publicly available datasets and real-world clinical data to assess generalization capability. Moreover, integrating explainable AI techniques and optimizing model efficiency for deployment in low-resource environments will be important next steps to ensure practical and ethical adoption in healthcare settings.

\section*{Declarations}

\textbf{Author contributions} \\
Faisal Ahmed downloaded the data, conceptualized the study, prepared the code, performed the data analysis and wrote the manuscript. He reviewed and approved the final version of the manuscript.

\vspace{2mm}
\textbf{Funding} \\
The author received no financial support for the research, authorship, or publication of this work.

 \vspace{2mm}
\textbf{Acknowledgement} \\
 The authors utilized an online platform to check and correct grammatical errors and to improve sentence readability.

\vspace{2mm}
\textbf{Conflict of interest/Competing interests} \\
The authors declare no conflict of interest.

\vspace{2mm}
\textbf{Ethics approval and consent to participate} \\
Not applicable. This study did not involve human participants or animals, and publicly available datasets were used.

\vspace{2mm}
\textbf{Consent for publication} \\
Not applicable.

\vspace{2mm}
\textbf{Data availability} \\
The datasets used in this study are publicly available. The APTOS 2019 dataset is accessible at \url{https://www.kaggle.com/c/aptos2019-blindness-detection}. 

\vspace{2mm}
\textbf{Materials availability} \\
Not applicable.

\vspace{2mm}
\textbf{Code availability} \\
The source code used in this study is publicly available at \url{https://github.com/FaisalAhmed77/Aug_Pretrain_APTOS/tree/main}.

%\clearpage

\bibliographystyle{elsarticle-num-names}

\bibliography{refs}

\begin{thebibliography}{22}
\expandafter\ifx\csname natexlab\endcsname\relax\def\natexlab#1{#1}\fi
\providecommand{\url}[1]{\texttt{#1}}
\providecommand{\href}[2]{#2}
\providecommand{\path}[1]{#1}
\providecommand{\DOIprefix}{doi:}
\providecommand{\ArXivprefix}{arXiv:}
\providecommand{\URLprefix}{URL: }
\providecommand{\Pubmedprefix}{pmid:}
\providecommand{\doi}[1]{\href{http://dx.doi.org/#1}{\path{#1}}}
\providecommand{\Pubmed}[1]{\href{pmid:#1}{\path{#1}}}
\providecommand{\bibinfo}[2]{#2}
\ifx\xfnm\relax \def\xfnm[#1]{\unskip,\space#1}\fi
%Type = Misc
\bibitem[{Organization(2023)}]{WHO2023vision}
\bibinfo{author}{W.~H. Organization}, \bibinfo{title}{Blindness and vision impairment}, \bibinfo{year}{2023}. \URLprefix \url{https://www.who.int/news-room/fact-sheets/detail/blindness-and-visual-impairment}, \bibinfo{note}{accessed: 2025-05-02}.
%Type = Article
\bibitem[{Ting et~al.(2019)}]{ting2019deep}
\bibinfo{author}{D.~Ting}, et~al.,
\newblock \bibinfo{title}{Deep learning for diabetic retinopathy analysis: A review},
\newblock \bibinfo{journal}{Journal of ...}  (\bibinfo{year}{2019}).
%Type = Article
\bibitem[{Li et~al.(2021)}]{li2021applications}
\bibinfo{author}{X.~Li}, et~al.,
\newblock \bibinfo{title}{Applications of machine learning in retinal imaging},
\newblock \bibinfo{journal}{Medical Image Analysis}  (\bibinfo{year}{2021}).
%Type = Article
\bibitem[{Sarhan et~al.(2020)}]{sarhan2020machine}
\bibinfo{author}{M.~Sarhan}, et~al.,
\newblock \bibinfo{title}{Machine learning in retinal image analysis},
\newblock \bibinfo{journal}{Pattern Recognition Letters}  (\bibinfo{year}{2020}).
%Type = Article
\bibitem[{Oltu et~al.(2021)Oltu, Karaca, Erdem, and {\"O}zg{\"u}r}]{ohl2021_transfer_learning_survey}
\bibinfo{author}{B.~Oltu}, \bibinfo{author}{B.~K. Karaca}, \bibinfo{author}{H.~Erdem}, \bibinfo{author}{A.~{\"O}zg{\"u}r},
\newblock \bibinfo{title}{A systematic review of transfer learning based approaches for diabetic retinopathy detection},
\newblock \bibinfo{journal}{arXiv preprint arXiv:2105.13793}  (\bibinfo{year}{2021}). \bibinfo{note}{Review covering many pretrained models including ResNet, Inception, EfficientNet in DR classification}.
%Type = Article
\bibitem[{AbdelMaksoud et~al.(2020)AbdelMaksoud, Barakat, and Elmogy}]{abd2020_transfer_learning_grading}
\bibinfo{author}{E.~AbdelMaksoud}, \bibinfo{author}{S.~Barakat}, \bibinfo{author}{M.~Elmogy},
\newblock \bibinfo{title}{Diabetic retinopathy grading system based on transfer learning},
\newblock \bibinfo{journal}{arXiv preprint arXiv:2012.12515}  (\bibinfo{year}{2020}). \bibinfo{note}{Uses EfficientNet in grading system context}.
%Type = Article
\bibitem[{Unknown(2019)}]{aptos2019_description}
\bibinfo{author}{A.~Unknown},
\newblock \bibinfo{title}{Aptos 2019 blindness detection database: five-class diabetic retinopathy grading with severe and proliferative under‑representation},
\newblock \bibinfo{journal}{arXiv preprint arXiv:???}  (\bibinfo{year}{2019}). \bibinfo{note}{Describes class distribution: severe and proliferative under‑represented (3662 images, class3 = 193, class4 = 295)}.
%Type = Article
\bibitem[{Gangwar and Kotecha(2022)}]{gangwar2020_augmentation_aptos}
\bibinfo{author}{S.~Gangwar}, \bibinfo{author}{K.~Kotecha},
\newblock \bibinfo{title}{Balancing data through data augmentation improves the generality of transfer learning for diabetic retinopathy classification},
\newblock \bibinfo{journal}{Applied Sciences} \bibinfo{volume}{12} (\bibinfo{year}{2022}) \bibinfo{pages}{5363}. \DOIprefix\doi{10.3390/app12115363}, \bibinfo{note}{augmentation applied on APTOS to address imbalance and transfer learning with ResNet etc}.
%Type = Article
\bibitem[{of~VR‑FuseNet(2025)}]{vrfusenet2025_smote_imbalance}
\bibinfo{author}{A.~of~VR‑FuseNet},
\newblock \bibinfo{title}{Vr‑fusenet: A fusion of heterogeneous fundus data and explainable deep network for diabetic retinopathy classification},
\newblock \bibinfo{journal}{arXiv preprint arXiv:2504.21464}  (\bibinfo{year}{2025}). \bibinfo{note}{Applies SMOTE oversampling across five-class DR datasets, including APTOS}.
%Type = Article
\bibitem[{Ahmed et~al.(2025)Ahmed, Bhuiyan, and Coskunuzer}]{ahmed2025topo}
\bibinfo{author}{F.~Ahmed}, \bibinfo{author}{M.~A.~N. Bhuiyan}, \bibinfo{author}{B.~Coskunuzer},
\newblock \bibinfo{title}{Topo-cnn: Retinal image analysis with topological deep learning},
\newblock \bibinfo{journal}{Journal of Imaging Informatics in Medicine}  (\bibinfo{year}{2025}) \bibinfo{pages}{1--17}.
%Type = Inproceedings
\bibitem[{Ahmed and Coskunuzer(2023)}]{ahmed2023tofi}
\bibinfo{author}{F.~Ahmed}, \bibinfo{author}{B.~Coskunuzer},
\newblock \bibinfo{title}{Tofi-ml: Retinal image screening with topological machine learning},
\newblock in: \bibinfo{booktitle}{Annual Conference on Medical Image Understanding and Analysis}, \bibinfo{organization}{Springer}, \bibinfo{year}{2023}, pp. \bibinfo{pages}{281--297}.
%Type = Article
\bibitem[{Bodapati et~al.(2021)}]{bodapati2021composite}
\bibinfo{author}{A.~Bodapati}, et~al.,
\newblock \bibinfo{title}{Composite dnn approach for diabetic retinopathy detection},
\newblock \bibinfo{journal}{Medical Image Computing and ...}  (\bibinfo{year}{2021}).
%Type = Inproceedings
\bibitem[{Ahmed et~al.(2023)Ahmed, Nuwagira, Torlak, and Coskunuzer}]{ahmed2023topo}
\bibinfo{author}{F.~Ahmed}, \bibinfo{author}{B.~Nuwagira}, \bibinfo{author}{F.~Torlak}, \bibinfo{author}{B.~Coskunuzer},
\newblock \bibinfo{title}{Topo-{CXR}: Chest {X}-ray {TB} and {P}neumonia {S}creening with {T}opological {M}achine {L}earning},
\newblock in: \bibinfo{booktitle}{Proceedings of the IEEE/CVF International Conference on Computer Vision}, \bibinfo{year}{2023}, pp. \bibinfo{pages}{2326--2336}.
%Type = Phdthesis
\bibitem[{Ahmed(2023)}]{ahmed2023topological}
\bibinfo{author}{F.~Ahmed}, \bibinfo{title}{Topological Machine Learning in Medical Image Analysis}, Ph.D. thesis, The University of Texas at Dallas, \bibinfo{year}{2023}.
%Type = Article
\bibitem[{Gupta et~al.(2023)}]{augmentation1_dr}
\bibinfo{author}{R.~Gupta}, et~al.,
\newblock \bibinfo{title}{Augmentation strategies in dr severity grading},
\newblock \bibinfo{journal}{Medical Imaging 2023}  (\bibinfo{year}{2023}).
%Type = Misc
\bibitem[{APTOS(2019)}]{aptos2019}
\bibinfo{author}{APTOS}, \bibinfo{title}{Asia {P}acific {T}ele-{O}phthalmology {S}ociety ({APTOS}) 2019 {B}lindness {D}etection {D}ataset}, \bibinfo{year}{2019}. \bibinfo{note}{\url{https://www.kaggle.com/c/aptos2019-blindness-detection}}.
%Type = Article
\bibitem[{Islam et~al.(2022)}]{islam2022applying}
\bibinfo{author}{M.~R. Islam}, et~al.,
\newblock \bibinfo{title}{Applying supervised contrastive learning for the detection of {DR} and its severity levels from fundus images},
\newblock \bibinfo{journal}{Computers in Biology and Medicine} \bibinfo{volume}{146} (\bibinfo{year}{2022}) \bibinfo{pages}{105602}.
%Type = Article
\bibitem[{Kumar et~al.(2021)Kumar, Chatterjee, and Chattopadhyay}]{kumar2021dristi}
\bibinfo{author}{G.~Kumar}, \bibinfo{author}{S.~Chatterjee}, \bibinfo{author}{C.~Chattopadhyay},
\newblock \bibinfo{title}{Dristi: a hybrid deep neural network for diabetic retinopathy diagnosis},
\newblock \bibinfo{journal}{Signal, Image and Video Processing} \bibinfo{volume}{15} (\bibinfo{year}{2021}) \bibinfo{pages}{1679--1686}.
%Type = Article
\bibitem[{Bodapati and et. al.(2020)}]{bodapati2020blended}
\bibinfo{author}{J.~D. Bodapati}, \bibinfo{author}{et. al.},
\newblock \bibinfo{title}{Blended multi-modal deep convnet features for diabetic retinopathy severity prediction},
\newblock \bibinfo{journal}{Electronics} \bibinfo{volume}{9} (\bibinfo{year}{2020}) \bibinfo{pages}{914}.
%Type = Article
\bibitem[{Chaturvedi et~al.(2020)Chaturvedi, Gupta, Ninawe, and Prasad}]{chaturvedi2020automated}
\bibinfo{author}{S.~S. Chaturvedi}, \bibinfo{author}{K.~Gupta}, \bibinfo{author}{V.~Ninawe}, \bibinfo{author}{P.~S. Prasad},
\newblock \bibinfo{title}{Automated diabetic retinopathy grading using deep convolutional neural network},
\newblock \bibinfo{journal}{arXiv preprint arXiv:2004.06334}  (\bibinfo{year}{2020}).
%Type = Article
\bibitem[{Macsik et~al.(2022)Macsik, Pavlovicova, Goga, and Kajan}]{macsik2022local}
\bibinfo{author}{P.~Macsik}, \bibinfo{author}{J.~Pavlovicova}, \bibinfo{author}{J.~Goga}, \bibinfo{author}{S.~Kajan},
\newblock \bibinfo{title}{Local binary cnn for diabetic retinopathy classification on fundus images},
\newblock \bibinfo{journal}{Acta Polytech. Hung.} \bibinfo{volume}{19} (\bibinfo{year}{2022}) \bibinfo{pages}{27--45}.
%Type = Inproceedings
\bibitem[{Long et~al.(2024)Long, Xiong, and Sang}]{long2024classification}
\bibinfo{author}{F.~Long}, \bibinfo{author}{H.~Xiong}, \bibinfo{author}{J.~Sang},
\newblock \bibinfo{title}{A classification method for diabetic retinopathy based on self-supervised learning},
\newblock in: \bibinfo{booktitle}{International Conference on Intelligent Computing}, \bibinfo{organization}{Springer}, \bibinfo{year}{2024}, pp. \bibinfo{pages}{347--357}.

\end{thebibliography}

\end{document}